\documentclass[conference]{IEEEtran}
\IEEEoverridecommandlockouts
\usepackage{cite}
\usepackage{amsmath,amssymb,amsfonts}
\usepackage{algorithmic}
\usepackage{graphicx}
\usepackage{textcomp}
\usepackage{xcolor}
\usepackage{url}
\usepackage{multirow}
\usepackage{booktabs}
\usepackage{listings}
\usepackage{xcolor}
\usepackage{fancyhdr}

\fancypagestyle{firstpage}{
  \fancyhf{}
  \fancyhead[L]{2024 First International Conference on Software, Systems and Information Technology (SSITCON)}
  \fancyfoot[C]{979-8-3503-5293-1/24/\$31.00~\copyright~2024 IEEE}
}
\def\BibTeX{{\rm B\kern-.05em{\sc i\kern-.025em b}\kern-.08em
    T\kern-.1667em\lower.7ex\hbox{E}\kern-.125emX}}
\begin{document}
\title{\textbf{LightFusionRec: Lightweight Transformers-Based Cross-Domain Recommendation Model} \\
}

\author{
  \IEEEauthorblockN{Vansh Kharidia\textsuperscript{*}, Dhruvi Paprunia\textsuperscript{*}, Prashasti Kanikar}
  \IEEEauthorblockA{
    Department of Computer Engineering,\\
    Mukesh Patel School of Technology Management \& Engineering,\\
    SVKM's Narsee Monjee Institute of Management Studies (NMIMS) Deemed-to-be University,\\
    Mumbai-400056, India\\
    Email: vansh.kharidia09@nmims.in, dhruvi.paprunia85@nmims.in, prashasti.kanikar@nmims.edu}
  \thanks{\textsuperscript{*}Equal contribution.}
}

\maketitle
\thispagestyle{firstpage}

\begin{abstract}
This paper presents LightFusionRec, a novel lightweight cross-domain recommendation system that integrates DistilBERT for textual feature extraction and FastText for genre embedding. Important issues in recommendation systems, such as data sparsity, computational efficiency, and cold start issues, are addressed in methodology.  LightFusionRec uses a small amount of information to produce precise and contextually relevant recommendations for many media formats by fusing genre vector embedding with natural language processing algorithms.  Tests conducted on extensive movie and book datasets show notable enhancements in suggestion quality when compared to conventional methods. Because of its lightweight design, the model can be used for a variety of purposes and allows for on-device inference. LightFusionRec is a noteworthy development in cross-domain recommendation systems, providing accurate and scalable recommendations to improve user experience on digital content platforms.

\end{abstract}

\section{\textbf{Introduction}}

The need for strong recommendation systems has increased due to the spread of digital material across many media channels. Conventional recommender systems typically suffer from the cold start issue, data sparsity, or merely a lack of context because they are compartmentalized inside a particular domain (reads, watches, etc.). New things and users are always adding to this problem, thus finding creative solutions that leverage information from other fields is essential.

To combat these challenges, LightFusionRec is introduced, which is a new cross-domain recommendation system to unify state-of-the-art NLP methods with generic text embedding. The method uses DistilBERT to encode semantically rich text features from content descriptions and the FastText model for genre feature extraction. The hybrid model connects different content domains (e.g., movies and books), but is a lightweight solution that runs in an efficient manner, making it specifically suitable for on-device inference.

\subsection{\textbf{Challenges Addressed}}
Current SOTA cross-domain recommendation models, especially when it comes to content recommendation, face some key challenges that limit their scope for widespread adoption and usefulness. The model addresses several of those key challenges: \newline

\subsubsection{{\textbf{Contextual understanding}}} 
\paragraph{\textbf{Challenge}} Existing systems often fail to capture the nuanced context of content descriptions and the semantic relationships between genres. \textsuperscript{~\cite{b28}}

\paragraph{\textbf{Solution}} By aligning DistilBERT-generated textual embedding with FastText-based genre vectors, LightFusionRec captures both the content description’s meaning and the semantic relationships between genres.
\newline
\subsubsection{\textbf{Getting cross-domain training data} }
\paragraph{\textbf{Challenge}} Traditional recommendation models require a lot of cross-domain training data for the same set of users, which is extremely difficult to procure.\textsuperscript{~\cite{b29}}

\paragraph{\textbf{Solution}} The model uses contextual understanding to recommend items that do not take any detailed user preferences or paired cross-domain data for a set of users while training.
\newline
\subsubsection{\textbf{Data Sparsity and Cold-Start Problem}}

\paragraph{\textbf{Challenge}} Difficulty in providing relevant recommendations to new users with a small interaction history. \textsuperscript{~\cite{b1}}

\paragraph{\textbf{Solution}} Traditional recommender systems often struggle with recommending content for new users or items due to a lack of interaction history. LightFusionRec efficiently  addresses the cold-start problem by focussing on these 2 parameters instead of user history:
\begin{enumerate}
    \item \textbf{Genre of Content}: This is represented by a genre vector (GV), which helps the model understand the types of movies or items that a user may prefer based on their genre preferences. For e-commerce, the genre of content refers to product categories.
    \item \textbf{Description Similarity}: The model uses vector embedding of content description to find niche interests, This means that if someone prefers science fiction, the embedding should be able to know which particular interests — like "Alien" movies.
\end{enumerate}

With the help of these two parameters, LightFusionRec can generate accurate content recommendations and does not necessarily need to first look for group viability or community interactions.
\newline
\subsubsection{\textbf{Scalability and efficiency} }
\paragraph{\textbf{Challenge}} Balancing recommendation quality with computational efficiency, especially for on-device applications. SOTA recommendation applications are compute-intensive and need to run in the cloud, which can be financially unsustainable.\textsuperscript{~\cite{b30}} 
\paragraph{\textbf{Solution}}Many traditional recommendation systems, especially those utilising deep learning, are computationally expensive and often require cloud infrastructure. LightFusionRec is designed to be lightweight, leveraging DistilBERT and precomputed genre embedding, allowing it to run efficiently on-device, offering scalability without compromising recommendation quality.

\section{\textbf{Related Works}}
Cross-domain recommendation systems have emerged as a major research area that overcomes the limitations of classic single-domain recommendation systems. In this study, a literature review is presented on the evolution of recommendation algorithms. Recent research has primarily focused on improving the performance of NLP techniques used in recommendation models, learning more comprehensive user- and item-embedding representations, implementing a co-training model to integrate multi-domain and multitask information, and utilizing deep Neural Network architectures in recommendation models.

\subsection{\textbf{Traditional Recommendation Systems}}\label{AA}

Traditional recommender systems are often divided into two categories: content-based filtering and collaborative filtering. The collaborative filtering approach predicts user preferences based on interactions between users and items, whereas content-based filtering makes recommendations based on item attributes. Resnick et al.'s (1994) \textsuperscript{\cite{b19}}  work on collaborative filtering utilizing social information filtering was one of the early works that impacted the present-day recommendation algorithms, although these systems suffer limitations like scalability issues and the cold start problem\cite{b19}. 

\subsection{\textbf{Advances in Deep Learning for Recommendations}}\label{AA}
The introduction of deep learning has resulted in a new revolution in recommender systems. He et al. (2017) \textsuperscript{\cite{b12}}introduced Neural Collaborative Filtering (NCF), which is the first model to include neural networks for learning user-item interactions. Variational encoders (VAE) have also been used in collaborative filtering; see Liang et al. (2018)\textsuperscript{\cite{b19}}, which may capture more complex context information for user items.

\subsection{\textbf{Cross-Domain Recommendation Systems}}\label{AA}
Cross-domain recommendation systems use data from several domains to boost the accuracy of their recommendations. Hu et al. (2009)\textsuperscript{\cite{b22}} conducted initial studies on the application of matrix factorization algorithms to transfer knowledge across disciplines. More recent approaches, like Man et al. (2017)\textsuperscript{\cite{b21}}, use deep learning models to collect latent properties across domains. Despite these advances, the challenge of successfully merging heterogeneous data from several fields remains.
\subsection{\textbf{NLP in Recommendations}}\label{AA}
The capability of recommender systems to understand and make use of textual material has been greatly enhanced by NLP approaches. Devlin et al. (2018) \textsuperscript{\cite{b23}}proved the effectiveness of transformers in extracting contextual information from text with BERT (Bidirectional Encoder Representations of Transformers). Newer models that offer efficient solutions appropriate for real-time applications include DistilBERT, a simplified version of BERT that was presented by Sanh et al. (2020)\textsuperscript{\cite{b24}}. To enhance content-based recommendations, these algorithms have been applied to extract rich textual information from item descriptions.

\subsection{\textbf{Genre embedding}}\label{AA}
The semantic connections among various genres are captured by genre embedding, which offer a concise comprehension of item properties. Word2Vec, first presented by Mikolov et al. (2013)\textsuperscript{\cite{b25}}, established the foundation for learning word representations. FastText is a Word2Vec enhancement created by Bojanowski et al. (2017)\textsuperscript{\cite{b27}} that takes sub-word information into account to further increase embedding quality. Recommendation algorithms have made better use of these embedding to comprehend genre semantics. Reddy et al. (2019) \textsuperscript{\cite{b6}} developed a movie recommendation technique based on genre correlations.

\subsection{\textbf{Hybrid Models for Cross-Domain Recommendations}}\label{AA}
Cross-domain recommendation research is leading the way with hybrid models that combine genre embedding and natural language processing approaches. By merging FastText-based genre vectors with text embedding produced by DistilBERT, the method expands upon this framework. By capturing contextual information from descriptions and semantic links between genres, this fusion overcomes the drawbacks of conventional models. Research has shown that hybrid techniques can effectively improve the precision of recommendations across different domains, as evidenced by the work of Wang et al. (2015)\textsuperscript{\cite{b26}} .

\subsection{\textbf{Recent Progress in Single-Domain and Cross-Domain Recommendations}}\label{AA}
Advancements in recent times have expanded the limits of cross-domain recommendation systems. Saraswat et al. (2018)\textsuperscript{\cite{b8}}, for instance, employed keyword and genre similarity to make cross-domain suggestions. Recently, further work has been done to address the cross-domain recommendation problem of cold start (Li et al. (2018)\textsuperscript{\cite{b1}}, Cao et al. (2022)\textsuperscript{\cite{b15}}). Transformers4Rec (Moreira et al., 2021)\textsuperscript{\cite{b2}} is one of the most well-known of the newly invented transformer structures specifically designed for recommendations; nevertheless, it has not yet been modified for cross-domain recommendations.

Li et al. (2023) \textsuperscript{\cite{b10}} provide Graph Transformers for Recommendations, a computationally costly model that requires vast amounts of data to execute, to demonstrate SOTA results in recommendation systems based on the transformers architecture. Using sentiment analysis for cross-domain recommendations, several cross-domain recommendation techniques, such Wang et al. (2020)\textsuperscript{\cite{b5}}, have taken use of current developments in NLP. The majority of large-scale recommendation systems in use today mostly rely on user history and data to provide recommendations. An increasing amount of research is being conducted on privacy-preserving techniques for recommendation systems as users grow more worried about their data and privacy. Federated learning is one method being used by researchers to make advantage of user data in recommendation models while protecting user privacy (Neumann et al., 2023)\textsuperscript{\cite{b7}}.

\section{\textbf{Proposed Method}}
\subsection{\textbf{Methodology}}

\begin{figure}
    \centering
    \includegraphics[width=1\linewidth]{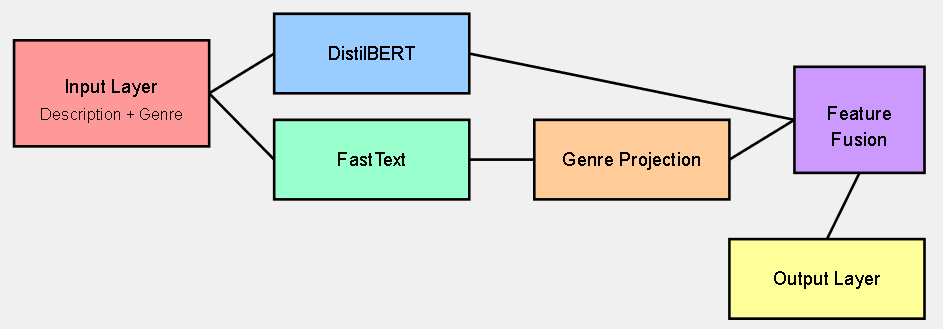}
    \caption{Architecture of LightFusionRec}
    \label{fig:enter-label}
\end{figure}
The architecture of LightFusionRec that was used to follow the methodology is shown in Fig 1. 

\subsubsection{\textbf{Inference Setup}}
\textbf{Software}: Python 3.7 or later \\
\textbf{Dependencies}: PyTorch 1.8.0+, Transformers 4.0.0+, FastText 0.9.2+, Scikit-Learn 0.24.0+, NumPy 1.20.0+, Pandas 1.2.0+ \\
\textbf{System Requirements:} CPU: Intel Core i3 or equivalent, Memory: 4 GB minimum (8 GB recommended), Storage: 512 MB free for model files and temporary data

The methodology flow is depicted in Fig. 2, which demonstrates how data was collected and prepared, then used DisitilBert for text feature extraction and FastText for genre vector preparation. Afterward, these steps were combined to train the reccomender model and precompute the book features. Assessment and Ablation Analysis Verify the Process's Accuracy. 

\subsubsection{\textbf{Data Collection and Preprocessing}}\label{AA}
The data sets used in the study comprise of movie and book descriptions, genres, and titles. The title is stored for output convenience, while the genres and description are used in model training. It was a conscious decision to use as minimal data as possible per record to ensure that the model is generalizable and performs well with minimal signifiers. The Amazon Reviews dataset is used for this paper.\textsuperscript{~\cite{b31}}. 

\begin{enumerate}

\item{\textbf{Data Cleaning}}:
Data cleaning was concluded by removing duplicate entries and removing entries with missing genre/description. The book dataset contained some entries where the title and/or description were not in English (some not even in a Latin script); Those entries were kept to ensure that the model does not break when it receives non-Latin input. After data cleaning, the dataset had ~45k books and  ~65k movies.

\item{\textbf{Genre Vector Preparation}} 
To capture the semantic relationships between genres, a FastText model was trained on the concatenated genres from both datasets. FastText, as proposed by Bojanowski et al. (2017)\textsuperscript{\cite{b27}}, extends Word2Vec by considering subword information, making it particularly suitable for learning embedding of short texts like genres. 

Let \( G \) be the set of all genres. For a genre \( g \in G \), its FastText embedding is denoted as:
\begin{equation}
v_g = \text{FastText}(g) \in \mathbb{R}^{50}
\end{equation}

Use of FastText compared to other embedding methods like Word2Vec or Glove as FastText takes into account sub-word information, while GloVe and Word2Vec treat words as atomic units.
FastText's approach leads to richer context representation, which helps the model distinguish between similar-sounding genres (e.g., "romance" and "romantic-comedy") by understanding subword differences.
Word2Vec and GloVe lack this level of contextual distinction, which makes it harder for them to distinguish between similar sounding genres like ‘romance’ and ‘romantic-comedy’.
\item {\textbf{Feature Fusion}} 
Feature Fusion is for alignment of textual features and genre vectors which affects the recommendation quality. It is done by the following ways:
\item{\textbf{Improved Content Understanding}}: Aligning textual features (e.g., keywords, descriptions) with genre vectors (categorical genre information) helps capture both semantic meaning and categorical preferences. This alignment allows for deeper insights into the nature of the content, which improves matching with user preferences.
\item{\textbf{Contextual Recommendations}}: Genre vectors have context because of their textual features. A work categorized as "science fiction," for instance, can also feature major textual themes like "AI" or "space travel." Aligning these vectors provides a more sophisticated understanding of content that is relevant to the user's context, which helps to improve genre-based suggestions.
\item{\textbf{Enhanced Personalization}}: Users may have a preference for certain genres, nevertheless within those genres, they may have different interests. (e.g. within the "romance" category, certain users may like narratives with "adventure" or "historical fiction" themes). The alignment captures both general and specialized preferences, aiding in the provision of more individualized recommendations.

 \end{enumerate}
\subsubsection{\textbf{Text Feature Extraction using DistilBERT}}\label{AA}

A lightweight variant of BERT called DistilBERT was utilized to extract rich textual features from content descriptions. Sanh et al. (2019)\textsuperscript{\cite{b24}} introduced DistilBERT, which is 40\% smaller and 60\% faster than BERT while maintaining 97\% of BERT performance.

For a content description \( d \), the DistilBERT embedding is 
\begin{equation}
    e_d = \text{DistilBERT}(d) \in \mathbb{R}^{768}
\end{equation}

\textbf{Tokenization and Encoding:} The DistilBERT tokenizer was used to tokenize and encode the descriptions in input IDs and attention masks, with a maximum sequence length of 128 tokens. The inputs which were encoded are then fed into the DistilBERT model to get the last hidden states, focusing on the [CLS] token for its summary representation of the sequence.

\subsubsection{\textbf{Cross Domain Feature Fusion}}\label{AA}
To integrate textual and genre information, a cross-domain recommender model was designed that combines the features from DistilBERT and the FastText genre vectors. The model architecture includes several key components:
\begin{enumerate}

\item{\textbf{Genre Projection}} The 50-dimensional genre vectors are projected to the 768-dimensional space to match the DistilBERT output.
the genre vector were projected to match the dimensionality of the text embedding:
\begin{equation}
    e_d = \text{DistilBERT}(d) \in \mathbb{R}^{768}
\end{equation}

where \( W_g \in \mathbb{R}^{768 \times 50} \) and \( b_g \in \mathbb{R}^{768} \) are learnable parameters.

\item{\textbf{Feature Concatenation and Fusion}} The textual and projected genre features are concatenated and passed through a fusion layer
The fused feature \( f \) is computed as:
\begin{equation}
    f = \text{ReLU}(W_f [e_d; p_g] + b_f)
\end{equation}
Where \( W_f\in \mathbb{R}^{768 \times 1536} \) and \( b_f\in \mathbb{R}^{768} \) are learnable parameters, and \( [e_d; p_g] \) denotes the concatenation of \( e_d \) and \( p_g \).

\end{enumerate}

\subsubsection{\textbf{Training the Recommender Model}}
The training process involves optimizing the model to minimize the loss of cosine embedding between the final features of movies and books.

\begin{enumerate}

\item {\textbf{Loss Function}} Cosine Embedding Loss was used in the model, which measures the similarity between the feature vectors.
Cosine embedding loss is used:
\begin{equation}
   L =
\begin{cases} 
1 - \text{sim}(m, b) & \text{if } y = 1 \\
\max(0, \text{sim}(m, b) - \text{margin}) & \text{if } y = -1
\end{cases} 
\end{equation}
Where \( y \) is 1 for positive pairs and -1 for negative pairs, and \(\text{margin}\) is a hyper-parameter.

\item{\textbf{Optimization}} The AdamW optimizer is used with a learning rate of $10^{-5} \times 2 \times 10^{-5}$ and a weight decay of 0.01. A cosine annealing learning rate scheduler with warm restarts was used to refine the learning rate during training.

\item{\textbf{Training Loop}} The model was trained for 2 epochs, using a batch size of 32. Gradients were clipped to a maximum norm of 1.0 to stabilize training. As the DistilBERT model along with FastText genre embedding already place similar content nearby, 2 epochs are enough to train the model as the purpose is to just tune it to the task. Training for longer, e.g. 10 epochs, resulted in the model over-fitting.

\end{enumerate}

\begin{figure}
    \centering
    \includegraphics[width=1 \linewidth]{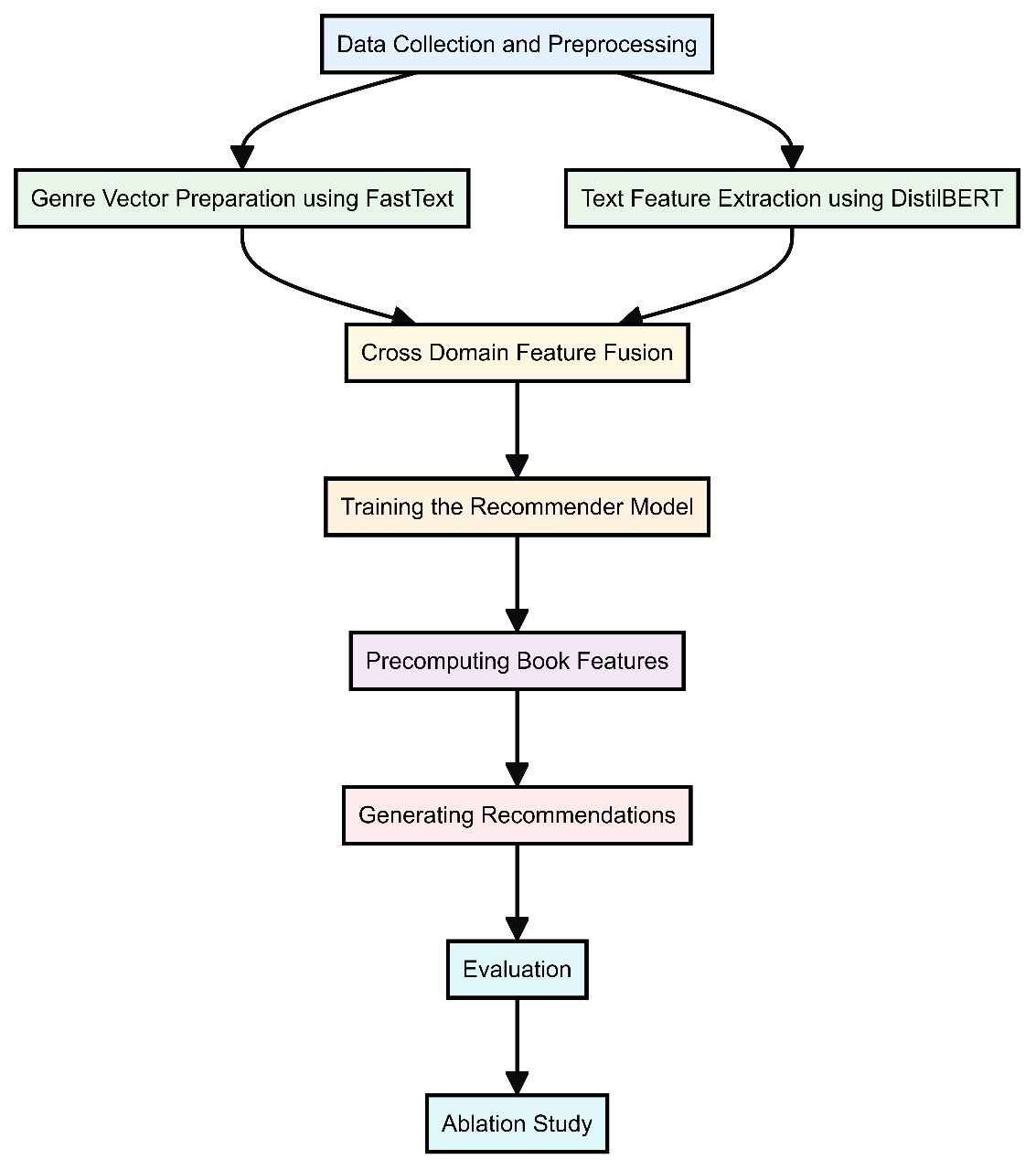}
    \caption{Methodology }
    \label{fig:enter-label}
\end{figure}

\subsubsection{\textbf{Pre-computing Book Features}}\label{AA}
To expedite the recommendation process, book features are pre-computed using the trained model. This is essential to speed up inference, as by pre-computing the books, the model only needs to compute the sample movies that the user provides during inference. This limited inference activity can be performed on devices which do not possess a lot of compute power.
\newline

\textbf{Batch Processing}: The books were processed in batches of 32 and their characteristics were extracted and stored for quick retrieval during the recommendation phase.
\begin{equation}
BF = \text{Model}(Book\;IDs, AM, GV)
\end{equation}
\noindent where BF is the Book Feature, AM the Attention Mask, and GV the Genre Vector.

\subsubsection{\textbf{Generating Recommendations}}\label{AA}
The recommendation process involves calculating the similarities between the combined movie features and the precomputed book features.

\begin{enumerate}

\item {\textbf{Movie Feature Extraction}}: For each input movie, features were extracted using the trained model and combined them to form a unified representation.
\begin{equation}
\text{Combined Movie Features} = \frac{1}{n} \sum_{i=1}^{n} \text{MF}_i
\end{equation}
\noindent where MF$_i$ denotes the $i$-th Movie Feature.

\item {\textbf{Similarity Calculation}}: The cosine similarity was calculated between the combined movie features and pre-calculated book features. In addition, the cosine similarities between genre and TF-IDF were also calculated and combined these metrics to generate final scores.
For a movie \( m \) and a book \( b \), their similarity is computed as:
\begin{equation}
    \text{sim}(m, b) = \text{cosine}(f_m, f_b)
\end{equation}

\item{\textbf{Top-K Recommendations}}: The top-K books with the highest combined scores were selected as recommendations.
\newline

\end{enumerate}

\subsection{\textbf{Pseudo-Code}}
The below pseudocode details the steps to train the LightFusionRec model including the optimization techniques that we've used like cosine annealing scheduler, AdamW optimizer with gradient clipping, etc. to help with replication of the model in different frameworks.
\begin{lstlisting}[
    language=Python,
    basicstyle=\ttfamily\footnotesize,
    numbers=left,
    numberstyle=\tiny,
    stepnumber=2,
    numbersep=5pt,
    backgroundcolor=\color{white},
    showspaces=false,
    showstringspaces=false,
    showtabs=false,
    tabsize=2,
    breaklines=true,
    breakatwhitespace=false,
    breakautoindent=true,
    captionpos=b
]
def train_LightFusionRec(dataset, epochs, batch_size, learning_rate):
    model = LightFusionRec()
    optimizer = AdamW(model.parameters(), lr=learning_rate)
    scheduler = CosineAnnealingWarmRestarts(optimizer, T_0=10, T_mult=2)

    for epoch in range(epochs):
        for batch in DataLoader(dataset, batch_size=batch_size):
            # Unpack batch
            descriptions, genres, labels = batch
            
            # Forward pass
            text_features = model.distilbert(descriptions)
            genre_features = model.fasttext(genres)
            projected_genres = model.genre_projection(genre_features)
            fused_features = model.feature_fusion(text_features, projected_genres)
            
            # Compute loss
            loss = cosine_embedding_loss(fused_features, labels)
            
            # Backward pass and optimization
            optimizer.zero_grad()
            loss.backward()
            torch.nn.utils.clip_grad_norm_(model.parameters(), max_norm=1.0)
            optimizer.step()

        scheduler.step()

    return model

# Main training loop
dataset = load_dataset()
model = train_LightFusionRec(dataset, epochs=2, batch_size=32, learning_rate=2e-5)
\end{lstlisting}

\subsection{\textbf{Evaluation}}
The model was evaluated using Root Mean Square Error (RMSE) and Mean Absolute Error (MAE), which are both effective metrics for assessing the accuracy of predictions. These metrics give the average size of errors in a set of predictions, allowing for a clear understanding of model performance across various thresholds. MAE and RMSE are calculated at the 20\%, 50\%, and 80\% thresholds to ensure a comprehensive evaluation of the performance of the model.

\subsubsection{\textbf{Evaluation Method}}\label{AA}
In the Amazon Reviews dataset, the list of users were extracted who had rated both movies and books. Then the books and movies were removed that the user had rated below 4 (out of 5), only keeping the books and movies that the users liked. The movies that each user liked were used to recommend books and based on those recommendations and the true values (the books that the user rated $\geq$ 4), The model was evaluated. The performance metrics that was used to measure are MAE, RMSE (Top 20\%, Top 50\%, and Top 80\%) was shown in Table 1.  

\begin{table}[h!]
\centering
\renewcommand{\arraystretch}{1.3} % Adjust row height
\begin{tabular}{|l|c|c|c|}
\hline
\textbf{Metric} & \textbf{Top 20\%} & \textbf{Top 50\%} & \textbf{Top 80\%} \\
\hline
MAE  & 0.6918 & 0.7710 & 0.8010 \\
\hline
RMSE & 0.6955 & 0.7039 & 0.7793 \\
\hline
\end{tabular}
\newline
\caption{LightFusionRec evaluation results}
\label{tab:metrics}
\end{table}

\subsubsection{\textbf{Ablation Study}}\label{AA}
Ablation studies were performed on the model, The model was evaluated using only the DistilBERT model, only using genre similarity and only using TF-IDF. The model significantly outperformed each of these models. The result is shown in Table 2 and visualised in Figure 3 which shows that the MAE and RMSE values of LightFusionRec is the least, indicating a good result. 

\begin{table}[h!]
\centering
\begin{tabular}{|l|l|c|c|c|}
\hline
\textbf{Model} & \textbf{Metric} & \textbf{Top 20\%} & \textbf{Top 50\%} & \textbf{Top 80\%} \\
\hline
\multirow{2}{*}{DistilBERT model} & MAE  & 1.9925 & 2.9920 & 3.254 \\
\cline{2-5}
& RMSE & 1.9961 & 2.6650 & 3.807 \\
\hline
\multirow{2}{*}{Genre-Based Similarity} & MAE  & 1.4259 & 2.1050 & 2.800 \\
\cline{2-5}
& RMSE & 1.6040 & 2.3615 & 2.894 \\
\hline
\multirow{2}{*}{tfidf\_only} & MAE  & 1.0328 & 1.2231 & 2.0175 \\
\cline{2-5}
& RMSE & 1.0513 & 1.4894 & 2.0426 \\
\hline
\end{tabular}
\newline
\caption{Results of Ablation Study}
\label{tab:combined_metrics}
\end{table}

\begin{figure}
    \centering
    \includegraphics[width=1\linewidth]{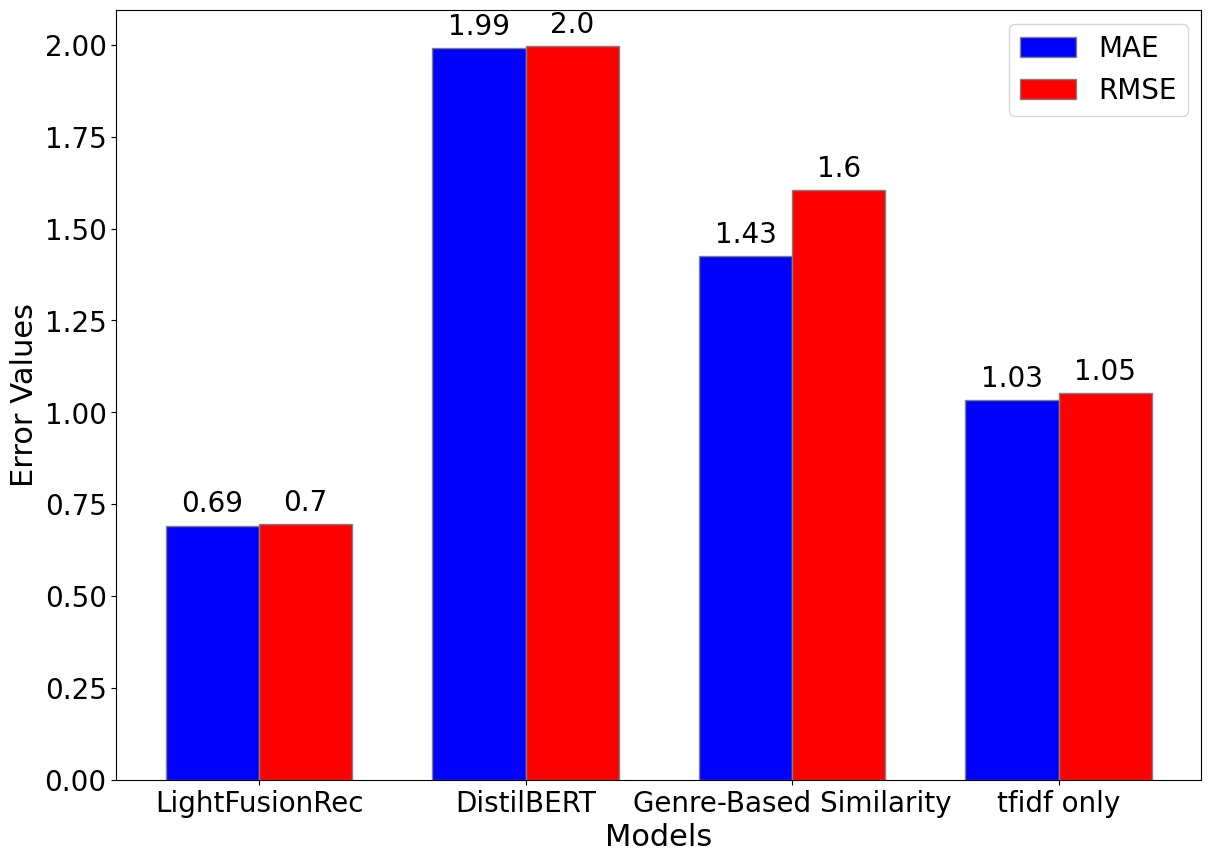}
    \caption{Comparing Model Results}
    \label{fig:enter-label}
\end{figure}

\subsection{\textbf{Interpretation of Results}}
The superior performance of LightFusionRec, particularly in cold-start scenarios, can be attributed to several key factors:

\begin{enumerate}
    \item \textbf{Rich Textual Understanding:} The use of DistilBERT allows for a nuanced interpretation of content descriptions, capturing semantic relationships that go beyond simple keyword matching.
    \item \textbf{Genre Semantics:} By incorporating FastText-based genre embedding, the model gains a deeper understanding of content categorization, which is particularly useful for cross-domain recommendations.
    \item \textbf{Feature Fusion:} The novel approach of combining textual and genre features allows the model to leverage complementary information sources, resulting in more robust recommendations.
   
\end{enumerate}

\subsection{\textbf{Comparison with Existing Approaches}}
LightFusionRec exhibits a number of benefits over current cross-domain recommendation systems, including:
\begin{enumerate}
    \item \textbf{Reduced Data Requirements}:  This approach can produce relevant recommendations with minimum input, in contrast to systems that require large amounts of user interaction data across different domains.
    \item \textbf{Improved Cold-Start Handling}: The model overcomes a major shortcoming of many current systems by producing pertinent recommendations with just one to three input pieces. 
    \item \textbf{Computational Efficiency}: A more lightweight model is produced by employing DistilBERT and effective genre projections as opposed to methods that make use of full-scale BERT or intricate neural architectures.
\end{enumerate}

\subsection{\textbf{Limitations and Potential Improvements}}
Although LightFusionRec displays encouraging outcomes, there are several things that should be done better:

\begin{enumerate}
    \item \textbf{Domain Specificity}: Books and movies are used to train the existing model. To validate its performance over a larger range of domains, more study is required. Subsequent work will test it for e-commerce and swap out the genre vectors for product category vectors.
    \item \textbf{User Personalization}: The present model emphasizes similarity based on content. Including user preference data may improve the personalization of recommendations.
    \item \textbf{Temporal Dynamics}: The model does not currently account for changes in user preferences or content popularity over time. Incorporating temporal features could be a valuable extension.
    \item \textbf{Needs Pre-computed Vectors of Resultant Domain}: When recommending books from movie/movies, pre-computed features of all the books that can be recommended.
\end{enumerate}

\section{\textbf{Conclusion}}
This paper introduced LightFusionRec, a novel lightweight cross-domain recommendation system that leverages advanced NLP techniques and genre embedding to provide accurate and contextually relevant recommendations across different media types.

\subsection{\textbf{Key Contributions}}

\begin{enumerate}
    \item \textbf{Novel hybrid architecture:} A unique model was introduced that integrates transformer-based language models with genre embedding, specifically designed for cross-domain content recommendation.
    \begin{itemize}
    \item \textbf{Efficient cross-domain feature fusion:} By combining DistilBERT for text embedding and FastText for genre embedding, the model captures rich semantic information across different content domains.
    \item \textbf{Lightweight architecture:} The use of DistilBERT and efficient genre projections allows for a compact model suitable for on-device inference.
    \item \textbf{Cold start mitigation:} The model generates relevant recommendations with minimal input (i.e. 1-3 items), hence addressing the cold start problem effectively .
\end{itemize} 
    \item \textbf{Feature fusion Technique}: The model uses an novel method to integrate textual and genre embedding, improving the capture of content similarity in different domains.
    \item \textbf{Cross-domain adaptation framework:} A flexible framework is offered that is adaptable to diverse content domains beyond movies and books, paving the way for future cross-domain recommendation systems. The same model can be tested for e-commerce, where genre vectors can be replaced with vectors representing product categories.
    \item \textbf{Lightweight Design:} By utilizing efficient embedding techniques, the model obtains high-quality recommendations while preserving suitable computational efficiency for on-device applications.

\end{enumerate}

The experimental results observed after applying the model on large-scale movie and book datasets demonstrate LightFusionRec's superior performance in generating high-quality cross-domain recommendations. This work underlines the potential of hybrid models in improving user experience on digital content platforms by offering extensive and accurate recommendations, thus addressing the integral limitations of traditional systems.

\subsection{\textbf{Implications}}\label{AA}
There are numerous significant ramifications for the recommendation system industry from LightFusionRec's success, including: 

\begin{enumerate}
\item \textbf{Cross-Domain Potential}:  It opens up new avenues for content discovery across various platforms by proving that it is feasible to use NLP and embedding approaches for cross-domain suggestions that work.
    \item\textbf{Efficiency in Recommendation}: The lightweight nature of the model illustrates that advanced, computationally intensive structures are not necessary to achieve high-quality recommendations. 
    \item \textbf{Cold-Start Solution}: The framework shows promise in tackling the enduring cold-start issue in recommendation systems through its performance with little input data.
\end{enumerate}

\section{Future Research Directions}

The following avenues are suggested for further investigation in light of the findings and LightFusionRec's present limitations:

\begin{enumerate}
    \item \textbf{Multi-Domain Expansion}: To evaluate the model's generalizability, expand its functionality to include a wider range of content domains, such as podcasts, e-commerce, music, and academic papers. 
    \item \textbf{Temporal Dynamics}: Look for ways to provide temporal information so the model can adjust over time to user preferences and content trends.
    \item \textbf{Privacy-Preserving Techniques}: Examine strategies for preserving or enhancing suggestion quality while reducing the requirement for private user information in line with the expanding privacy concerns. 

\end{enumerate}
Future research can build on LightFusionRec's foundation by tackling these areas, which will result in more adaptable, efficient, and successful cross-domain recommendation systems that can greatly improve user experiences on a variety of digital platforms.

\end{document}